# Investigating Responsible AI for Scientific Research: An Empirical Study

Muneera Bano, Didar Zowghi, *Member, IEEE,* Pip Shea, Georgina Ibarra

*Abstract*—Scientific research organizations that are developing and deploying Artificial Intelligence (AI) systems are at the intersection of technological progress and ethical considerations. The push for Responsible AI (RAI) in such institutions underscores the increasing emphasis on integrating ethical considerations within AI design and development, championing core values like fairness, accountability, and transparency. For scientific research organizations, prioritizing these practices is paramount not just for mitigating biases and ensuring inclusivity, but also for fostering trust in AI systems among both users and broader stakeholders. In this paper, we explore the practices at a research organization concerning RAI practices, aiming to assess the awareness and preparedness regarding the ethical risks inherent in AI design and development. We have adopted a mixed-method research approach, utilising a comprehensive survey combined with follow-up in-depth interviews with selected participants from AI-related projects. Our results have revealed certain knowledge gaps concerning ethical, responsible, and inclusive AI, with limitations in awareness of the available AI ethics frameworks. This revealed an overarching underestimation of the ethical risks that AI technologies can present, especially when implemented without proper guidelines and governance. Our findings reveal the need for a holistic and multi-tiered strategy to uplift capabilities and better support science research teams for responsible, ethical, and inclusive AI development and deployment.

*Impact Statement* — This paper's examination of responsible AI practices underscores the potential implications of ethically guided AI for scientific research. The research findings can inform future research to ensure more reliable, unbiased scientific outcomes, strengthening the credibility and impact of research findings. Embracing responsible and ethical AI practices can significantly enhance public trust in AI technologies, fostering broader acceptance and support for AI-centric scientific endeavors. Leadership in this area could serve as a model for global research institutions, catalyzing a shift towards more ethical and socially responsive AI applications in science. The adoption of responsible AI practices can drive economic growth by fostering innovation, enhancing efficiency, and creating new market opportunities in the technology sector.

## I. INTRODUCTION

Responsible and ethical AI refers to the development and deployment of Artificial Intelligence (AI) systems that adhere to principles of fairness, accountability, transparency, and inclusivity [1]. Responsible AI (RAI) focuses on minimizing potential risks, biases, and negative consequences while maximizing the benefits to society [2]. Ethical AI emphasizes the moral principles guiding the design and use of AI systems, ensuring they do not perpetuate discrimination, infringe on privacy, or undermine human dignity [3]. Both responsible and ethical AI foster trust among users and stakeholders and are essential for the sustainable and equitable advancement of AI technology.

Numerous guidelines and principles have emerged for responsible and ethical AI, but there remains a significant gap in their practical implementations [3-5]. Inconsistencies in interpreting and applying these principles, coupled with a lack of diverse perspectives, raise concerns about their effectiveness [6, 7]. Operationalizing responsible and ethical AI principles is challenging due to the absence of proven methods, common professional norms, and robust legal accountability mechanisms [8-10]. Moreover, guidelines often overlook the practical aspects of business practices and political economies surrounding AI systems, leading to issues such as ethics washing, corporate secrecy, and competitive norms [4].

Operationalizing responsible and ethical AI guidelines involves translating abstract principles into tangible actions, processes, and practices that can be integrated into integrated into an organisation's systems, tools and workflows [10, 11]. Effective operationalization leads to greater transparency and trust among users and stakeholders, bridging the gap between theoretical ideals and real-world implementation [12].

This exploratory study focuses on understanding the practices of responsible and ethical AI within the research organization, Commonwealth Scientific and Industrial Research Organisation (CSIRO)[1]. CSIRO is Australia's national science research agency. It is one of the largest and most diverse research organizations in the world, renowned for its groundbreaking work in various fields such as agriculture,

Muneera Bano, PhD, is a senior research scientist at CSIRO, Australia. She is member of research team on Responsible AI and AI Diversity and Inclusion. Email: muneera.bano@csiro.au

Professor Didar Zowghi, PhD, is a senior principal research scientist, and Team lead of AI Diversity and Inclusion at CSIRO, Australia. Email: didar.zowghi@csiro.au

Pip Shea, PhD is a senior product designer and team lead at CSIRO, Australia. Email: pip.shea@csiro.au

Georgina Ibarra is a product and design lead at CSIRO, Australia. Email: georgina.ibarra@csiro.au

[1] https://www.csiro.au/

climate science, energy, health, and information technology. CSIRO plays a crucial role in advancing scientific knowledge and innovation, contributing significantly to Australia's economic development and solving some of the most pressing global challenges. Its commitment to collaborative research has positioned it as a leader in scientific discovery and application, making substantial impacts both locally and internationally.

As the national science agency, CSIRO is expected to perform science and innovation in a responsible way; therefore, science teams need better ways to be intentional and explicit about the steps they are taking to create responsible and ethical AI. In another recently published report by CSIRO, public perceptions of responsible innovation were captured as a benchmark study of 4000 Australians[2]. Their analysis surfaced that "risk management practices increase the belief that socially responsible outcomes will be delivered"[3]. This offers additional evidence of the value of our project's inquiry into risk assessment, mitigation, and management.

Our study is timely, considering the research organisations are already being challenged to deliver increasingly ethical AI [13]. The report offers evidence of widespread AI adoption across multiple science domains. It also outlines several key areas for science and research organisations to uplift their AI capabilities. The area of uplift most significant to this study is the rise of ethical expectations and regulations in AI for Science. Research organisations are already being challenged to deliver increasingly ethical AI. These pressures are coming from emerging regulations and societal expectations, the momentum that is set to continue as "the AI ethics performance bar is likely to be higher and more tightly regulated into the future"[4].

This study utilizes a mixed-method approach through a survey questionnaire and interview. The primary objective is to uncover gaps in awareness and understanding of responsible, ethical, and inclusive AI, and a review of current practices to ensure responsible AI development and deployment within CSIRO in future. Our participant cohort included CSIRO research scientists, project managers, executive managers, software engineers, and user experience designers. The selected AI projects (called AI for Missions)[5] within CSIRO are multidisciplinary, applied research initiatives designed to tackle large-scale systemic issues such as eliminating plastic waste, drought resilience, and biosecurity.

The remainder of this paper is structured as follows: Section 3 offers background information and motivation for our research. Section 4 delves into the details of our research methodology. Section 5 presents the results obtained from the survey and interviews. Section 6 discusses the results, while Section 7 concludes the paper and suggests directions for future research.

## II. BACKGROUND AND MOTIVATION

The growing interest in RAI has led to an increasing body of literature exploring the practical challenges and ethical issues involved in implementing responsible AI in organizational contexts. One aspect of this literature is the examination of how organizational culture and structures impact the effectiveness of RAI initiatives in practice [9]. Eitel-Porter [8] argues that businesses require strong, mandated governance controls, including tools for managing processes and creating associated audit trails to enforce their principles. Benjamins et al. [12] further discuss the practical case of a large organization that is putting in place a company-wide methodology to minimize the risk of undesired consequences of AI. The importance of practical methodologies and the need for empirical insights into how organizations deal with AI ethics are also emphasized in the literature [14, 15]. Morley, [16] argues that existing translational tools and methods are either too flexible (and thus vulnerable to 'ethics washing') or too strict (unresponsive to context), suggesting the need for alternative approaches. Lu et al. [1] propose a Responsible AI Pattern Catalogue based on a Multivocal Literature Review, which provides systematic and actionable guidance for stakeholders to implement responsible AI from a system perspective.

The growing importance of ethical and responsible AI development has prompted various governments and organizations to take significant initiatives in promoting and implementing ethical AI practices such as the World Economic Forum (WEF) [17] and the Australian Government's Ethical AI Framework [18]. Despite the ongoing efforts, a critical review of the literature, as well as documents on ethical AI frameworks, reveals a gap in understanding how organizations grasp and apply ethical and responsible AI concepts in practice. A recent report by the Human Technology Institute[6] found that critical decisions shaping the design, development, and procurement of AI technologies in large corporations "*are heavily influenced by a single 'guru', perceived as the organizational expert in AI'*"[7.] Furthermore, some organisations reported that AI-related risks were not being managed holistically as a part of the IT governance systems. Of additional concern was that some AI systems had been "*approved by legal and compliance teams with little to no practical knowledge of AI systems*".

Implementing responsible and ethical principles in AI remains challenging [10] due to the absence of proven methods, common professional norms, and robust legal accountability mechanisms [4]. While AI ethics guidelines often focus on algorithmic decision-making, they tend to overlook the practical and operational aspects of the business practices and political economies surrounding AI systems [19]. This oversight can lead to issues such as "ethics washing", corporate secrecy, and competitive and speculative norms [4]. The event of "internal revolt" at Google and the firing of AI ethics researchers further exemplify how ethical AI frameworks can be rendered ineffective due to business decision-making and economic logic [20].

---

[2] https://publications.csiro.au/rpr/download?pid=csiro:EP2022-0408&dsid=DS1
[3] https://www.uts.edu.au/human-technology-institute/news/report-launch-state-ai-governance-australia
[4] https://www.csiro.au/en/research/technology-space/ai/artificial-intelligence-for-science-report
[5] https://research.csiro.au/ai4m/
[6] https://www.uts.edu.au/media/563631
[7] https://www.uts.edu.au/sites/default/files/2023-05/HTI%20The%20State%20of%20AI%20Governance%20in%20Australia%20-%2031%20May%202023.pdf

There is ongoing fierce debate on the effectiveness and practical applicability of AI ethical guidelines. Munn [6] highlights the ineffectiveness of current AI ethical principles in mitigating racial, social, and environmental harms associated with AI technologies, primarily due to their contested nature, isolation, and lack of enforceability. In response, Lundgren [21] argues that AI ethics should be viewed as trade-offs, enabling guidelines to provide action guidance through explicit choices and communication. Despite conceptual disagreements, Lundgren suggests building on existing frameworks, focusing on areas of agreement, and setting clear requirements for data protection and fairness measures.

It is crucial to survey the level of understanding of ethical and responsible AI concepts within an organization, as it can inform tailored strategies, governance models, and training programs that address specific needs and challenges. Our survey-driven approach, therefore, holds significant potential for CSIRO in enhancing the effectiveness of responsible and ethical AI initiatives and fostering a culture across various industries to address challenges of bias [22], fairness [23], transparency [24] and, explainable and responsible AI [25].

## III. RESEARCH METHODOLOGY

We have employed a mixed-method approach in our exploratory research. Our survey participant cohort included CSIRO research scientists, project managers, executive managers, software engineers, and user experience designers. Whereas our interview participant cohort was all involved in a program of work known internally as AI for Missions[8]. Missions within CSIRO are large multidisciplinary initiatives focussed around six challenges with a national focus: 1) Health and wellbeing 2) Food security and quality 3) Secure Australia and region 4) Resilient and valuable environments 5) Sustainable energy and resources, and 6) Future industries. The AI for Missions program specifically funds internal projects to develop AI systems that work towards these national challenges. Data was collected for both studies in second half of 2022.

### A. SURVEY

The survey was designed with the objective of identifying projects that are using AI to further their objectives in scientific research, the roles involved in designing, developing, and deploying AI systems, and the process followed in the implementation of AI systems or components. The survey also aimed to examine the extent to which these projects were aware of and adhered to the some of the existing AI Ethics frameworks (such as [17, 18]) in their design and development of AI systems.

The survey was designed with 39 questions [See Appendix A], covering different aspects of AI implementation such as demographics, project context, inclusive and ethical AI, risk assessment, and others. Participation in the survey was not mandatory, and the number of responses varied for each question, with a minimum of 23 and a maximum of 35. The analysis of the survey data was conducted in two steps. In the first step, the survey questionnaire was categorized into themes based on the nature of the questions. The categories were developed in consultation with the team members and included project context, inclusive and ethical AI, risk assessment, and others. The aim was to group the questions thematically, which would facilitate cross-examination of the responses in the second step of the analysis. In the second step, insights were drawn from individual questions and aggregated for each category. For close-ended questions, frequency analysis was applied to the responses to determine the most common answers. For open-ended questions, thematic analysis was applied to find recurring themes in the answers.

The results presented in this report should not be considered a definitive study on the use of AI within CSIRO but rather a snapshot of the responses from a relatively small number of individuals within the organization who consented to participate in the survey. One notable limitation is the low response rate due to voluntary participation. Despite these limitations, the information provided offers insights into key areas for future improvement, serving as a starting point for further exploration and analysis.

### B. Interview

The primary objectives of the interviews were to:

- Conduct contextual inquiries to deepen our understanding of the AI project's context within CSIRO.
- Identify and delve into potential ethical risks associated with the AI systems.
- Gain insights to guide the design of future AI solutions that integrate responsible, ethical, and inclusive AI principles in design, development, deployment and governance.

The interview participants encompassed a diverse range of projects utilizing AI, offering a comprehensive overview of AI's applications across various scientific domains. These projects spanned disciplines such as manufacturing, causal analysis, drug discovery, health monitoring, ingredient enhancement, earth observation, energy system management, animal health assessment, supply chain oversight, and advanced imaging. Each project presented unique challenges, employing AI in distinct ways to address specific issues within their respective domains.

The in-depth interviews conducted with team members from these projects yielded valuable insights into the multifaceted use of AI in diverse scientific contexts. The discussions delved into the AI techniques, algorithms, and models employed within each project. Interviewees shared their experiences, outlining encountered hurdles, and innovative remedies devised. Furthermore, they provided nuanced perspectives on the impact of AI on their work and industries, highlighting both the advantageous and detrimental aspects, alongside associated risks. Participants were chosen through purposive sampling, based on their expertise and participation in the designated projects. All individuals who consented to be part of the interview study were included, making up a sample of 28 individuals.

---

[8] https://research.csiro.au/ai4m/

Interviews with each participant were carried out virtually and video-recorded to ensure precision. The duration of these interviews spanned from 60 to 90 minutes. To ensure comprehensive coverage and consistency across the interviews, a structured questionnaire of 20 questions was designed and utilized. This questionnaire is detailed in Appendix B. With the aid of Dovetail's live transcription services[9], all interviews were transcribed verbatim, guaranteeing conformity to the participants' authentic expressions and sentiments.

Post-transcription, dedicated efforts were directed towards de-identifying and anonymizing the interview transcripts, eliminating any reference to participants or specific project details before any analysis. This step was imperative to safeguard the confidentiality of participants and uphold the integrity of the study as recommended by the Ethics approval for the study by CSIRO [10].

In the initial thematic analysis, the interviewer (also a 3rd co-author) manually coded the interviews for themes using the Dovetail platform, focusing on capturing the depth and nuances of the participant responses. In parallel, GPT-4 was deployed to discern emerging themes and to generate more themes that may have been missed in manual analysis.

Following this, a comprehensive comparison of the themes identified by GPT-4 and the manual coding was conducted. This comparison involved extensive discussions held over three meetings between the researchers. During these sessions, the codes and themes generated by both methods were juxtaposed, evaluated for alignment and divergence, and deliberated upon. The aim was to refine and converge on the final themes that best represented the data. The collaborative nature of this process, where both AI-generated and human-coded themes were scrutinized side by side, added an enhanced layer of rigor to the research. The active participation of two additional authors (2nd and 3rd) in these discussions ensured that the resulting themes were not only comprehensive but also well-grounded in the context of each AI project, leading to more nuanced and profound interpretations.

In the concluding phase, themes were examined across the diverse project landscapes. This comparative approach enabled the research team to identify recurring patterns as well as variances among the projects. It further deepened the understanding of how these themes played out in varying contexts. Through this inter-project analysis, eight unique themes centred around Ethical and Responsible AI emerged. These themes shed light on the intricate challenges and considerations when deploying AI in scientific endeavours. The structured, sequential methodology laid the foundation for a robust investigation into the experiences and viewpoints of participants. Merging AI-driven insights with human analysis, this study promises a multifaceted perspective on the role of AI in scientific research.

## IV. RESULTS

### A. Survey

#### 1) Projects

The survey offers valuable insights into the AI-based projects being undertaken by respondents within CSIRO. Firstly, the results indicate that there is a diverse range of AI projects being worked on by the respondents. This underscores the broad scope and applicability of AI technology in various scientific domains. Secondly, the study found that customer interviews were the dominant activity utilized for building a shared understanding of AI development. Thirdly, people were found to be predominantly involved in advisory roles within AI systems. This validated our assumption that due to the lack of tools to operationalise Responsible AI, the stakeholders who remain most invested in its implementation are those accountable for the high-level risks and opportunities afforded by AI technologies. Lastly, more than 60% of the projects (out of 25 responses) were found to be currently under development in CSIRO, with over 50% of these projects utilizing historical data. These findings provide important insights for the development of science-specific Responsible AI tools. Knowing that historical data sets are so heavily relied upon by scientists may help internal data managers and data stewards to prioritise these data sets to better foreground Responsible AI in data processes. Furthermore, this preference for historical data surfaces an opportunity for CSIRO to direct resources to support RAI processes that underpin other data sources.

#### 2) Inclusive AI

The survey findings indicate that there is still room for improvement in this regard. Firstly, 40% of the respondents reported having no understanding of Inclusive AI, while 48% of the respondents did not think their AI systems were inclusive. Secondly, 68% of the respondents either did not know or did not think that their data needed to be more representative. This may also reflect that some scientific data might only account for plants and animals — and not humans — therefore, science teams might be less likely to prioritise ethical approaches to AI development.

#### 3) Ethical AI

The findings indicate that there is a need for greater awareness of and support to surface ethical issues in these scientific fields. Firstly, 70% of the respondents reported not having used any AI Ethics Framework, with 17% of respondents indicating they were not even aware of its existence. This highlights the need for increased awareness and promotion of ethical frameworks to ensure that AI-based projects are developed in a socially responsible and ethical manner. Secondly, 83% of the respondents reported not having used any additional resources on ethical AI design or development. A small percentage of respondents conflated the AI ethics framework with the Human & Animal Research Ethics policies. This suggests a lack of emphasis on ethical considerations in the development and implementation of AI-based projects.

---

[9] https://dovetail.com/

[10] The demographic details of the participants of the survey and interview, as well as the projects are not provided for compliance with Ethics Approval from CSIRO.

*4) Risk Assessment*

The findings highlight the need for increased emphasis on risk assessment and ethical considerations in the development and deployment of AI systems. Firstly, the study found that 70% of respondents had not used any AI Ethics Framework, with 17% indicating they were not even aware of its existence. Secondly, close to 40% of respondents were not aware of any ethical risks associated with their AI projects, and 40% were not using any risk assessment techniques, with 20% indicating that they were not aware of any risk assessment within their projects.

We found that privacy protection and information security were identified as the top risk factors within AI projects, based on the responses. However, fairness, bias, accountability, and explainable AI, which are usually considered to be significant risks for AI systems, were not identified as top risks in the responses. Finally, almost 48% of the projects had not undergone any risk assessment, which underscores the need for greater emphasis on the importance of risk assessment in the development and implementation of AI-based projects.

*5) Data*

The results indicate that 68% of the respondents either did not know or did not think their data needed to be more representative. This suggests that there may be a need for greater awareness and understanding of the importance of representative data in AI systems to ensure fairness and reduce bias. We observed that not many respondents were working on "missing data", which has been considered a top source of uncertainty that can impact the validity of AI decision-making. Additionally, We found that bias detection, explainable AI, and accountability were not considered the main priorities in addressing ethical risks in AI, according to the responses.

*6) Humans*

The results of the study reveal that AI is being used as an assistant or advisor rather than a replacement for human labor. Out of 25 responses, 76% of the projects are not replacing any human activity, indicating that humans are still playing a significant role in AI projects. In projects that involve human-AI collaboration, 32% are classified as "Advisory", meaning that AI presents options to humans for decision-making, and requires some input from people. Additionally, 24% of the projects are classified as "Assistive", which means that AI assists human decision-making by performing basic processes, requiring significant input by people.

*7) Process*

In terms of process, the results indicate that the dominant part of AI projects are related to model training or the development of new machine learning (ML) algorithms. Additionally, it was found that the ML approach that dominates the process is "Black Box" rather than "Explainable". The use of explainable AI (XAI) [26] techniques in the development process could help to address this limitation, providing greater transparency and insights into how these models make their predictions or decisions.

*8) Summary of survey results*

The results indicate that while many AI projects at CSIRO are contributing to the organization's strategies, there is room for improvement in several areas. For instance, a proportion of respondents do not fully understand or prioritize inclusive AI. Some projects have not undergone risk assessment and have not used any AI Ethics Framework or additional resources on ethical AI design and development. In terms of process, while model training and developing new ML algorithms dominate, there is a need for more attention to be paid to areas such as missing data, and bias detection. These findings highlight the importance of increased awareness and action around ethical and inclusive AI practices, as well as rigorous risk assessment and attention to process.

*B. Interview*

*1) Ethical and Responsible AI*

Regarding Ethical and Responsible AI, various themes have emerged from interviews of participants from different project teams. A predominant reluctance was observed towards conducting risk assessments for AI systems due to reasons ranging from perceived lack of necessity to concerns about slowing research progress. To mitigate these concerns, teams were encouraged to shift their perspectives from viewing ethical considerations as liabilities to seeing them as value-adding opportunities. Such a transition from a problem-focused to a value-centric approach serves to operationalize responsible AI, stimulating project momentum.

Participants highlighted the challenges of managing potential biases, ensuring fairness, protecting privacy, and understanding the broader societal impacts of AI technologies. For instance, in one project focused on developing an AI system for animal health inspection, participants voiced concerns about possible biases and fairness implications, particularly concerning smaller entities that might be unable to afford such technology. This perspective is mirrored in the statement*: "Balancing explainability with reliability and safety can often be a challenge"*. Some participants felt that the current ethical guidelines and principles were rather broad and doubted their direct relevance to their specific projects, indicating a desire for more tailored guidelines. *"The AI ethics principles seem very overarching and general. We have similar guidelines in research, like the national research code of conduct"*. These participants also confessed to having a limited grasp of AI ethical principles, hinting at a potential knowledge gap in the domain: *"I've only glanced at the principles, so my understanding is quite preliminary"*. This sentiment was echoed by other participants: *"I've only briefly reviewed them"*.

The analysis highlights four projects where ***ensuring fairness and representation*** is integral to responsible AI

systems within the agriculture and food sectors. In one project, the team understands the necessity of including a diverse range of seed growers in their training data. By doing so, they aim to avoid an unfair advantage to well-resourced farms in limited geographical areas, ensuring that their AI system is fair and representative. The second case emphasizes the centrality of fairness in developing traceability verification systems across the food supply chain. Researchers on the project, using AI Ethical Principles [17, 18], have created a framework for evaluating the use of these systems to ensure benefits are distributed fairly from breeders to consumers.

Another theme delves into the critical **need for transparency and explainability in AI systems**, particularly in sectors related to agriculture, environment, and food supply. One project dealing with livestock disease testing emphasizes the legal necessity of having a comprehensible AI system. Should their testing processes come under scrutiny, they aim to elucidate how the results were obtained and how AI was instrumental in the process. Similarly, for water quality forecasting, there is a significant need for transparency concerning the inherent uncertainties of forecasts. This transparency is vital not only for managing user expectations but also due to the open nature of data repositories, where control over data use and potential misuse is limited. AI systems need to be managed with care to ensure their usage does not inadvertently lead to negative impacts.

*Accountability* pertains to an AI project team's realization that monitoring and fine-tuning an AI system proves more challenging than their physics-based models, due to the 'black box' nature of AI. This situation underscores the imperative role of human intervention in AI processes, emphasizing the necessity for identifiable and accountable individuals to oversee the outcomes produced by these AI systems.

*2) Risk Assessment and Strategic Management:*

Risk assessment and strategic management emerged as consistent motifs throughout the interviews, underscoring the significance of pinpointing and mitigating risks in AI projects. In one project, the inception of a risk register was highlighted, signifying a methodical tactic towards project oversight and risk deterrence. Participants from another project accentuated the imperative of ceaseless surveillance of AI systems. This was to pinpoint inaccuracies, biases, or unforeseen consequences, thereby reiterating the ethos of conscientious AI practices. Meanwhile, some participants expressed scepticism about the efficacy of current risk assessment procedures, viewing them more as procedural impediments than valuable assets. This sentiment is encapsulated in the statement, *"Understand how useful the internal risk assessment processes are, and mostly they're not useful for people. They're just an admin burden."*

Our analysis of interview participants working on various AI projects revealed the complexities inherent in the ethical risk assessment of AI systems. In one project, AI was utilized in farming to balance environmental advantages and potential biodiversity risks. While this project displayed remarkable environmental benefits, it also highlighted the potential for biodiversity loss. Lastly, a water quality forecasting project, aimed at societal and environmental well-being, underscored the need for clear differentiation between project objectives and ethical AI practices to prevent false security and unanticipated consequences. These cases underline the importance of mindful integration of AI ethics in projects serving public good. According to some of the participants, strategic communication and tangible ethical actions play a crucial role in shaping public perception and ensuring responsible AI deployment. They underscore the potential of AI to contribute to societal and environmental well-being when used responsibly, emphasizing the need for tools that effectively demonstrate these contributions and ethical behaviours in AI practices.

*3) Managing Data and Addressing Uncertainty:*

Data management and the intricacies of navigating uncertainty consistently emerged as key themes throughout the interviews. Participants emphasized the pivotal role of various stages in the data lifecycle, from collection and pre-processing to integration, and how data uncertainties can profoundly affect AI system reliability. In one project, participants underscored the indispensable value of premium-quality data, shedding light on the repercussions of having missing or incomplete datasets on AI models. A resonating sentiment was, *"Your model's effectiveness is directly proportional to the quality of data it's built on."* In another project, a team member delved into data uncertainty, noting, *"The accuracy of measurements is paramount... there's always this element of unpredictability when contrasting our simulation models against real-world scenarios."*

The nature and management of different data types also prominently featured in the discussions. Participants shared insights into the utilization of diverse data sources, such as historical records, satellite earth observations, and hyperspectral imaging, within their projects. The meticulous processes of data management – encompassing collection, cleansing, pre-processing, and integration – were spotlighted as key stages to ensure the highest standards of data quality and reliability.

*4) Collaboration and Engaging Stakeholders:*

Collaboration and engaging stakeholders were observed as important themes across various interviews. Given the inherently interdisciplinary character of AI projects, participants underscored the crucial role of collaboration for domain experts, forging ties with industry partners, and maintaining ongoing involvement of key stakeholders.

In one project, the shared knowledge emanating from collaborative activities was spotlighted as a key factor for project success. Another project delved into the nuances of managing stakeholder perspectives and potential apprehensions, especially when rolling out innovative technologies in sectors that are traditionally more resistant to change. A participant from this project observed, *"As I delineate our value propositions and engage in dialogues with stakeholders, I am constantly on the lookout for diverse perspectives. Every piece of input and varied viewpoint on the subject is invaluable."*

*5) Adaptive Strategies and Societal Impact in AI*

Another theme spotlighted two intertwined facets in project of AI for science, the imperative for adaptability in AI projects and the overarching societal implications of AI assimilation. Embracing the fluidity of the AI landscape, participants highlighted the necessity of evolving projects to cope with fluctuating conditions, underscoring the essence of quickly adaptable strategies. Beyond immediate project adaptability, there was a conspicuous acknowledgment of the broader ramifications of AI integration. The analysis further highlighted the importance of comprehending both immediate and protracted consequences, encompassing potential risks, stakeholder viewpoints, and potential socio-economic ripples. This lays emphasis on the overarching need for a holistic societal perspective in AI undertakings, aiming to ensure safety, consistency, and public confidence.

*6) Regulatory Compliance and Ethical Guidelines:*

The discussions underscored two cardinal principles: the significance of regulatory compliance and the imperative of ethical stewardship in AI ventures. Participants emphasized the importance of synchronizing with AI ethical principles [17, 18], ensuring that initiatives are anchored in ethical benchmarks. Such a focus underscores a resolute commitment to conscientious AI methodologies.

*7) Democratization of AI Technology*

Participants voiced concerns over potential imbalances in AI adoption, noting that initial costs might predominantly favour major entities, thereby fostering inequalities. It was unanimously affirmed that conscious strides must be taken to guarantee equitable AI technology dissemination, forestalling the amplification of existing disparities.

*8) Summary*

The discussions within AI for scientific research project teams highlight a multi-faceted and complex situation of challenges and considerations. Central themes included ethical responsibilities, the dual-edged sword of risk assessment, data management intricacies, the imperative of collaboration, and the broader societal ramifications of AI projects. A consistent emphasis was placed on regulatory compliance and ensuring equitable access to AI technologies.

## V. Discussion

*A. Synthesis of results from Survey and Interview*

The two studies provide important insights into the state of ethical and responsible AI practices within AI projects at CSIRO. Both reports identify key challenges and insights within responsible and inclusive AI development and deployment.

*B. Challenges*

The survey report reveals that while AI projects contribute significantly to CSIRO's strategies, gaps exist in understanding and prioritizing inclusive AI, conducting thorough risk assessments, and employing ethical AI resources [17, 18]. The survey also highlights an overemphasis on model training and developing new machine learning algorithms, at the expense of addressing missing data and bias detection. Similarly, the interview study report underscores the need to shift perceptions of ethical considerations in AI from being viewed as liabilities to potential opportunities. It explores several key themes of responsible and ethical AI that, collectively, call for comprehensive ethical risk assessments, the incorporation of human values, fair practices, privacy, security, reliability, transparency, contestability, and accountability in AI systems.

*C. Comparison to existing literature*

Our results can be compared to existing literature on the operationalization of responsible and ethical AI. The findings of the survey highlight several areas where CSIRO can benefit from improvements in operationalizing responsible and ethical AI. For example, the low awareness and utilization of AI ethical frameworks and risk assessment techniques and their operationalisation are consistent with previous research [3, 4, 6, 10, 27, 28] that has highlighted a lack of practical guidance on integrating responsible and ethical AI principles into AI systems [11, 12]. Similarly, the emphasis on responsible, ethical and inclusive AI principles aligns with previous research that has called for a more robust approach to ensuring future AI systems do not perpetuate biases or discrimination. The survey's finding that many AI projects have not been subjected to rigorous risk assessments echoes Eitel-Porter's [8] argument for the necessity of strong governance controls, also highlighted by WEF and Gov frameworks [17, 18]. The limited engagement with AI Ethics Frameworks among respondents at CSIRO is another important observation, further underscoring the literature's emphasis on the disconnect between the prolific development of ethical AI principles and their real-world enactment [2].

*D. Areas of improvements*

The survey findings, combined with insights from the interview study, uncover several areas where enhancements can be made for a more robust practice of responsible and ethical AI within CSIRO. First and foremost, there exists a pressing requirement for heightened **awareness** and advocacy of ethical frameworks, including the AI Ethics Frameworks [17, 18], and risk assessment methodologies. This will ensure that AI-oriented initiatives undergo rigorous scrutiny and are structured in a manner that adheres to ethical and responsible principles. Additionally, an amplified focus on **rules and regulations** to foster responsible, ethical and inclusive AI principles is essential to guarantee the development of AI systems that uphold fairness and representation, ultimately creating more equitable outcomes. **Effective communication** and collaboration with stakeholders also need amplification to align AI-based projects with organizational objectives and societal norms. The interviews further brought to light the necessity of adopting a **human-centric approach to AI ethics**, endorsing the creation of AI systems that are not only user-friendly but also trustworthy. Moreover, reframing ethical considerations in AI from a liability to a value proposition was suggested as a crucial strategy to operationalize responsible AI. Lastly, integrating strategic actions that demonstrate the contributions of ethical AI to societal and environmental well-being can play a pivotal role in shaping public perception and facilitating responsible AI deployment.

*E. Trade-off analysis*

The findings of both the survey and the interview study suggest that a balancing act may exist between the pragmatic execution of AI-based projects and the ethical contemplations that form their foundation. For instance, the infrequent use of ethical frameworks, including the AI Ethics Frameworks, and risk assessment methodologies, as evidenced by the survey, are indicated to be the result of conflicting resource demands or the need for swift progress in AI-based project development. Likewise, the lack of awareness of inclusive AI principles might stem from a deficiency of expertise or insufficient time for training in this area. The interview study revelations concur with survey findings, which highlight the need to transition from viewing ethical considerations as liabilities of time and resources to recognising them as potential value-additions for the future. Together, these findings underscore the necessity of finding an equilibrium between the practical and ethical considerations of AI-based projects to ensure their effectiveness, responsibility, and alignment with societal values.

*F. Future of AI for Science*

The rapid integration of AI, particularly Generative AI and Large Language Models (LLMs), into scientific research ushers in a transformative era with expanded options for AI co-pilots in research [29, 30]. This evolution, however, brings to the forefront the urgent need for a deeper understanding and heightened awareness of responsible and ethical AI usage. As highlighted by recent studies, there is an imperative need for organizations in any field to adopt a core set of AI ethics frameworks. This includes thorough ethical risk assessments, fairness, transparency, contestability, and accountability.

Yet, the application of these ethics in AI is not a one-size-fits-all scenario. Every organization must undertake a thoughtful, introspective process to adapt these principles to their unique circumstances. This customization considers factors such as the nature of the AI project, the stakeholders involved, and the broader societal implications. Adapting to a tailored ethical AI approach presents its challenges but is crucial. It necessitates a significant shift in business process modeling and reengineering to inherently incorporate ethical AI principles. This shift ensures that AI solutions are not only innovative but also in harmony with societal values and norms. This evolution, while potentially disruptive in the short term, can fortify an organization's position in the long run, ensuring resilience, public trust, and sustained innovation in a world increasingly leaning on AI.

VI. Conclusion and Future Work

In conclusion, the present study provides valuable insights into the practices of responsible and ethical AI within the CSIRO. The findings highlight the need for increased emphasis on the awareness, understanding, and implementation of ethical, responsible and inclusive AI principles in the development and deployment of AI systems within scientific research. The results of the study provide a basis for further research on the operationalization of responsible and ethical AI in the scientific community and contribute to the ongoing discussion on the role of AI in shaping our future.

References

1. Lu, Q., et al., *Responsible AI Pattern Catalogue: a Multivocal Literature Review.* arXiv preprint arXiv:2209.04963, 2022.
2. Ghallab, M., *Responsible AI: requirements and challenges.* AI Perspectives, 2019. **1**(1): p. 1-7.
3. Jobin, A., M. Ienca, and E. Vayena, *The global landscape of AI ethics guidelines.* Nature machine intelligence, 2019. **1**(9): p. 389-399.
4. Attard-Frost, B., A. De los Ríos, and D.R. Walters, *The ethics of AI business practices: a review of 47 AI ethics guidelines.* AI and Ethics, 2022: p. 1-18.
5. Zowghi, D. and F. da Rimini, *Diversity and Inclusion in Artificial Intelligence.* arXiv preprint arXiv:2305.12728, 2023.
6. Munn, L., *The uselessness of AI ethics.* AI and Ethics, 2023. **3**(3): p. 869-877.
7. Shams, R.A., D. Zowghi, and M. Bano, *AI and the quest for diversity and inclusion: a systematic literature review.* AI and Ethics, 2023: p. 1-28.
8. Eitel-Porter, R., *Beyond the promise: implementing ethical AI.* AI and Ethics, 2021. **1**: p. 73-80.
9. Rakova, B., et al., *Where responsible AI meets reality: Practitioner perspectives on enablers for shifting organizational practices.* Proceedings of the ACM on Human-Computer Interaction, 2021. **5**(CSCW1): p. 1-23.
10. Bano, M., et al., *AI for All: Operationalising Diversity and Inclusion Requirements for AI Systems.* arXiv preprint arXiv:2311.14695, 2023.
11. Zhu, L., et al., *AI and ethics—Operationalizing responsible AI.* Humanity Driven AI: Productivity, Well-being, Sustainability and Partnership, 2022: p. 15-33.
12. Benjamins, R., A. Barbado, and D. Sierra, *Responsible AI by design in practice.* arXiv preprint arXiv:1909.12838, 2019.
13. Hajkowicz, S., et al., *Artificial intelligence for science–Adoption trends and future development pathways.* Brisbane, Australia, 2022.
14. Stahl, B.C., et al., *Organisational responses to the ethical issues of artificial intelligence.* AI & SOCIETY, 2022. **37**(1): p. 23-37.
15. Ryan, M., et al., *Research and practice of AI ethics: a case study approach juxtaposing academic discourse with organisational reality.* Science and Engineering Ethics, 2021. **27**: p. 1-29.
16. Morley, J., et al., *Ethics as a service: a pragmatic operationalisation of AI ethics.* Minds and Machines, 2021. **31**(2): p. 239-256.
17. World-Economic-Forum. *AI Ethics Framework.* 2023 [cited 2023 10 May]; Available from: https://www.weforum.org/projects/ai-ethics-framework.
18. Australian-Government. *Australia's Artificial Intelligence Ethics Framework.* 2019 [cited 2023 10th May ]; Available from: https://www.industry.gov.au/publications/australias-artificial-intelligence-ethics-framework.
19. Mittelstadt, B., *Principles alone cannot guarantee ethical AI.* Nature machine intelligence, 2019. **1**(11): p. 501-507.
20. Dastin, J. and P. Dave. *Exclusive: Google pledges changes to research oversight after internal revolt*. 2021 [cited 2023 10th May]; Available from: https://www.reuters.com/article/us-alphabet-google-research-exclusive/exclusive-google-pledges-changes-to-research-oversight-after-internal-revolt-idUSKBN2AP1AC.
21. Lundgren, B., *In defense of ethical guidelines.* AI and Ethics, 2023: p. 1-8.
22. Leavy, S., B. O'Sullivan, and E. Siapera, *Data, power and bias in artificial intelligence.* arXiv preprint arXiv:2008.07341, 2020.
23. Xivuri, K. and H. Twinomurinzi. *A systematic review of fairness in artificial intelligence algorithms.* in *Responsible AI and Analytics for an Ethical and Inclusive Digitized Society: 20th IFIP WG 6.11 Conference on e-Business, e-Services and e-Society, I3E 2021,*

*Galway, Ireland, September 1–3, 2021, Proceedings 20*. 2021. Springer.
24. Felzmann, H., et al., *Towards transparency by design for artificial intelligence.* Science and Engineering Ethics, 2020. **26**(6): p. 3333-3361.
25. Arrieta, A.B., et al., *Explainable Artificial Intelligence (XAI): Concepts, taxonomies, opportunities and challenges toward responsible AI.* Information fusion, 2020. **58**: p. 82-115.
26. Angelov, P.P., et al., *Explainable artificial intelligence: an analytical review.* Wiley Interdisciplinary Reviews: Data Mining and Knowledge Discovery, 2021. **11**(5): p. e1424.
27. Hagendorff, T., *The ethics of AI ethics: An evaluation of guidelines.* Minds and machines, 2020. **30**(1): p. 99-120.
28. Cachat-Rosset, G. and A. Klarsfeld, *Diversity, Equity, and Inclusion in Artificial Intelligence: An Evaluation of Guidelines.* Applied Artificial Intelligence, 2023. **37**(1): p. 2176618.
29. Bano, M., D. Zowghi, and J. Whittle, *Exploring Qualitative Research Using LLMs.* arXiv preprint arXiv:2306.13298, 2023.
30. Bano, M., et al., *Large Language Models for Qualitative Research in Software Engineering: Exploring Opportunities and Challenges.* 2023.

# APPENDIX A – SURVEY QUESTIONNAIRE

A. *Participant's Demographic Questions*
 1. How many projects have you worked on at CSIRO that utilise AI/ML?
 2. Of these projects, have any been involved with a CSIRO's strategies?
 3. Which of the following best describes your role in the project?
 4. Which level were you when you worked on this AI/ML project?

B. *Project Context Questions*
 5. What is the purpose of the project where AI is used?
 6. Is there a partner or customer investing in the AI solution?
 7. Which industry sector and/or market segment would you identify this partner or customer with?
 8. Which business area best identifies the work of this partner or customer?
 9. Are you involved in designing or developing the technology for the AI system/component?

C. *Project related Questions*
 10. Briefly describe the problem that the AI system/component will help to solve?
 11. Who specifically experiences this problem?
 12. What types of AI system/component is/was developed to help solve this problem?
 13. What activities have you done to build shared understanding about the problem you are trying to solve using AI (both internally and with the partner or customer)?
 14. Who are the users of this AI system or component?
 15. What is the relationship between people and the AI system in your project? [Humans]
 16. Is your project's AI system/component?
 17. What type of data does your project use for the AI system or component you are developing?

D. *Inclusive AI*
 18. What is your understanding of "inclusive" AI systems?
 19. Do you think your AI system is inclusive?
 20. Which of the groups your dataset is representative of?
 21. Do you think your data needs to be more representative? [Question 29]

E. *Uncertainty*
 22. What sources of uncertainty exist in the data that could impact the validity/success of your AI system?
 23. Is the AI system/component intended to replace humans in the activities? [Humans]
 24. Do you intend to report on uncertainties evolving from data and models used in the AI system?
 25. How might uncertainty influence quality and reliability of AI decisions and trust in systems?

F. *Humans*
  [Question 15 & 23]

G. *Risk Assessment*
 26. Are you aware of ethical risks in your AI system/component?
 27. Are you using any techniques to manage these risks?
  [Question 29 & 35]

H. *Data*
  [Questions 20 & 21]
 28. What data management processes you are involved in, as related to the design or development of the AI system or component?
 29. What techniques you or your team use to reduce ethical risk in your AI system or component?

I. *Process*
 30. Which of the algorithm and model development processes you are involved in, as related to the design or development of the AI system or component?
 31. What is your ML approach?

J. *Ethical AI*
 32. How often do you monitor the output of your AI system for inaccuracies, biases, or behaviour that is not intended?
 33. Have you utilised available AI Ethics Frameworks for guidance on ethical AI principles?
 34. Do you use any frameworks or guidelines that help you to apply AI ethics in the design and development and deployment of your AI system/ component?
 35. Has your project undergone any kind of risk analysis or assessment process?
 36. Did your project use any of the following approaches to facilitate responsible/ethical AI?
 37. If you selected any of the above, briefly describe your involvement in the process (if any).
 38. What do you think is obstructing the adoption of methods that help projects adhere to the ethical principles?

K. *Follow up Question*
  Thank you for your participation in our survey. Your experiences are a valuable contribution to our investigation of the ethical issues permeating AI/ML. Are you interested in being contacted for a follow up interview?

# APPENDIX B – INTERVIEW QUESTIONNAIRE

**Q1. What are your roles on the AI project?**
**Q2. Which industry sector is your AI project targeted at?**
 A. AI in Health (includes aged care and disability services)
 B. AI in Mining
 C. AI in Law
 D. AI in Finance (includes insurance and superannuation)
 E. AI in Agribusiness (includes natural resources and environment management)
 F. AI in Cyber Security
 G. AI in Education
 H. AI in Defence
 I. AI in Infrastructure (includes transportation, energy and water services, telecommunications, waste management and smart cities)
 J. AI in Manufacturing
 K. AI in R&D or Innovation (includes any ground-breaking, new and emerging AI process or technology in additional fields not mentioned in the above industry sectors)
 L. AI in Environment
 M. Other

**Q3. Which business areas does your AI project target?**
 A. Accounting and finance
 B. Customer service
 C. Human resources
 D. IT
 E. Legal, risk and compliance
 F. Supply chain
 G. Marketing
 H. Operations

    I. Manufacturing
    J. Research and development
    K. Sales
    L. Strategy
    M. other

**Q4. What types of AI system are you developing?**
    A. Recognition systems
    B. Language processing
    C. Automated decision making
    D. Recommender systems
    E. Computer vision
    F. Other

**Q5. Can you describe the problem the AI system will help to solve? - Who specifically experiences this problem?**

**Q6. Who are the key stakeholders of this AI system?**

**Q7. Who are the end users of this AI system?**

**Q8. What is the relationship between people and the AI system in your project?**
    A. *Assistive* (provides information) — AI assists human decision making by performing basic processes, requiring significant input by people.
    B. *Advisory* (presents options) — AI augments human decision making by performing complex processes, requiring some input by people.
    C. *Autonomous* (makes decisions without human direction) — the machine operates independently, making decisions with real world impacts and requiring no input by people.
    D. *Discovery* (doing a novel or new task, eg. In pattern recognition that could not be achieved by a human alone)
    E. other

**Q9. Can you describe an example of a Responsible AI practice?**

**Q10 -** *<<omitted for anonymising the paper>>*

**Q11. How has your project assessed and analysed the ethical risks associated with the AI system? - Did you use a risk analysis methodology to do this assessment?**

**Q12. What type of data does your project use for the AI system or component you are developing?**
    A. Historical data
    B. Synthetic data
    C. Other

**Q13. What type of data management processes are you involved in?**
    **A.** Data collection
    **B.** Data cleaning
    **C.** Data processing
    **D.** Data analysis
    **E.** Data integration
    **F.** Data governance
    **G.** Data design (eg. Database design)
    **H.** Merging of data sources
    **I.** Dealing with hybrid data (eg. From more than one source or type)
    **J.** Dealing with missing data
    **K.** Other

**Q14. What sources of uncertainty exist in the data that could impact the success of your AI system?**
    A. Missing data
    B. Accuracy of measurements
    C. Inconsistent data
    D. Incomplete data
    E. Biased data
    F. Other

**Q15. How does uncertainty influence quality and reliability of AI decisions?**

**Q16. How will you monitor the output of your AI system for inaccuracies, biases, or outcomes that are not intended?**

**Q17. Have you used or do you plan to use the CSIRO Research Data Planner?**

**Q18. Have you used or do you plan to use the CSIRO Enterprise Risk Consequence table?**

**Q19. Do you think you sufficiently understand the AI Ethical Principles to do a self-assessment of your AI/ML project?**

**Q20. How do you think Responsible AI practices could help your project achieve its desired objectives and research impact?**

**Author's Biographies**

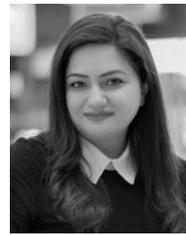

**1st Author: Muneera Bano, PhD** is Senior Research Scientist and member of Diversity and Inclusion team at CSIRO's Data61. She is an award-winning scholar, is passionate advocate for gender equity in STEM. She is a Diversity Inclusion and Belongingness (DIB) officer at Data61 and a member of the 'Equity, Diversity and Inclusion' committee for Science and Technology Australia. Muneera graduated with a PhD in Software Engineering from UTS in 2015. She has published more than 50 research articles in notable international forums on Software Engineering. Her research, influenced by her interest in AI and Diversity and Inclusion, emphasizes human-centric technologies.

Contact her at muneera.bano@csiro.au
Official Webpage: https://people.csiro.au/B/M/muneera-bano

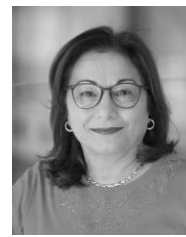

**2nd Author: Didar Zowghi, (PhD, IEEE Member since 1995)** is a Senior Principal Research Scientist and leads the science team for Diversity and Inclusion in AI at CSIRO's Data61. She is an Emeritus Professor at the University of Technology Sydney (UTS) and conjoint professor at the University of New South Wales (UNSW). She has decades of experience in Software Engineering research and practice. In 2019 she received the IEEE Lifetime Service Award for her contributions to the research community, and in 2022, the Distinguished Educator Award from IEEE Computer Society TCSE. She has published over 220 research articles in prestigious conferences and journals and has co-authored papers with over 100 researchers from 30+ countries.

Contact her at didar.zowghi@csiro.au
Official Webpage: https://people.csiro.au/z/D/Didar-Zowghi

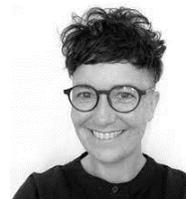

**3rd Author: Pip Shea, PhD** is a senior Product Designer and Team Lead at CSIRO's Data61. Currently working to transform and scale scientific processes using AI, robotics, and other emerging technologies within Data61's Science Digital program. Her experience of design practice spans 25 years and is shaped by periods as an academic researcher, artist, and community development practitioner. She has a PhD in Media & Communication from the Queensland University of Technology (2014). Contact her at pip.shea@csiro.au

**4th Author: Georgina Ibarra,** is product design lead at CSIRO's Data61. Contact her at georgina.ibarra@csiro.au